\documentclass{article}
\pdfoutput=1

\usepackage[preprint,nonatbib]{neurips_2019}

\usepackage[utf8]{inputenc} %
\usepackage[T1]{fontenc}    %
\usepackage{booktabs}       %
\usepackage{nicefrac}       %
\usepackage{microtype}      %

\usepackage{amssymb}   %
\usepackage{amsmath}   %
\usepackage{amsfonts}  %
\usepackage{amsthm}    %
\usepackage{latexsym}  %
\usepackage{stmaryrd}  %

\usepackage{graphicx}  %
\usepackage{color}     %
\usepackage{xcolor}    %

\usepackage{verbatim}  %
\usepackage{listings}  %
\usepackage{enumitem}  %
\usepackage{algorithm} %
\usepackage[noend]{algpseudocode} %

\usepackage{url}       %
\usepackage{xspace}    %
\usepackage{etoolbox}  %
\usepackage{hyperref}  %

\usepackage{array}     %
\usepackage{multirow}  %
\usepackage{multicol}  %
\usepackage{hhline}    %
\usepackage{dcolumn}   %

\usepackage{graphicx}
\usepackage{subcaption}
\usepackage{mwe}
\usepackage{amssymb}   %
\usepackage{pifont}    %

\usepackage{tikz}      %
\usepackage{pgfplots}  %
\pgfplotsset{compat=1.14}

\newcommand\refsec[1]{Section~\ref{sec:#1}}

\newcommand\reffig[1]{Figure~\ref{fig:#1}}

\newcommand\reftab[1]{Table~\ref{tab:#1}}

\ifthenelse{\isundefined{\definition}}{\newtheorem{definition}{Definition}}{}
\ifthenelse{\isundefined{\assumption}}{\newtheorem{assumption}{Assumption}}{}
\ifthenelse{\isundefined{\hypothesis}}{\newtheorem{hypothesis}{Hypothesis}}{}
\ifthenelse{\isundefined{\proposition}}{\newtheorem{proposition}{Proposition}}{}
\ifthenelse{\isundefined{\theorem}}{\newtheorem{theorem}{Theorem}}{}
\ifthenelse{\isundefined{\lemma}}{\newtheorem{lemma}{Lemma}}{}
\ifthenelse{\isundefined{\corollary}}{\newtheorem{corollary}{Corollary}}{}
\ifthenelse{\isundefined{\alg}}{\newtheorem{alg}{Algorithm}}{}
\ifthenelse{\isundefined{\example}}{\newtheorem{example}{Example}}{}

\definecolor{applegreen}{rgb}{0.55, 0.71, 0.0}

\newcommand{\pseu}[1]{#1}

\newcommand{\mtg}{\textsc{MTG}}

\newcommand{\hearthstoneshort}{\textsc{HS}}
\newcommand{\django}{\textsc{Django}}

\newcommand{\concode}{\textsc{CONCODE}}
\newcommand{\naps}{\textsc{NAPS}}
\newcommand{\ours}{SPoC}

\newcommand{\cmark}{\textcolor{applegreen}{\ding{51}}}%
\newcommand{\xmark}{\textcolor{red}{\ding{55}}}%

\newcommand{\tin}{T^\mathrm{in}}
\newcommand{\tout}{T^\mathrm{out}}
\newcommand{\htin}{\tilde T^\mathrm{in}}
\newcommand{\htout}{\tilde T^\mathrm{out}}
\newcommand{\err}{\mathrm{err}}

\newcommand{\testp}{\textsc{TestP}\xspace}
\newcommand{\testw}{\textsc{TestW}\xspace}

\title{SPoC: Search-based Pseudocode to Code}

\author{%
 Sumith Kulal\thanks{Equal contributions.}
\ \ ,\  Panupong Pasupat$^{\ast}$, \ Kartik Chandra,\  Mina Lee, \\
 {\bf Oded Padon,\  Alex Aiken,\  Percy Liang}
  \\
 Department of Computer Science \\
 Stanford University\\
 \texttt{\{sumith,ppasupat,kach,minalee,padon,aaiken,pliang\}@cs.stanford.edu} \\
}

\begin{document}

\maketitle

\begin{abstract}

We consider the task of mapping %
pseudocode
to long programs that are functionally correct.
Given test cases as a mechanism to validate programs,
we search over the space of
possible translations of the pseudocode
to find a program that
passes the validation.
However, without proper credit assignment
to localize the sources of program failures,
it is difficult to guide search %
toward more promising programs.
We propose to perform credit assignment
based on signals from compilation errors,
which constitute 88.7\% of program failures.
Concretely,
we treat the translation of each pseudocode line
as a discrete portion of the program,
and whenever a synthesized program fails to compile,
an error localization method
tries to identify the portion
of the program responsible for the failure.
We then focus search over alternative translations
of the pseudocode for those portions.
For evaluation,
we collected the SPoC dataset (Search-based Pseudocode to Code)
containing 18,356 programs with human-authored pseudocode
and test cases.
Under a budget of 100 program compilations,
performing search improves the synthesis success rate
over using the top-one translation of the pseudocode
from 25.6\%
to 44.7\%.

\end{abstract}

\section{Introduction}

We consider the task of mapping natural language descriptions to functionally correct computer programs that are long enough
to have significant intermediate state (e.g., 10–20 lines) and
perform non-trivial computations.
Previous work on executable semantic parsing
mainly focuses on translating short text descriptions
to a one-line program
\cite{zelle96geoquery,vadas2005programming,zettlemoyer07relaxed,zhong2017seq2sql,lin2018nl2bash,dong2018coarse},
and while recent work explored
generating longer programs from text descriptions
\cite{ling2016latent,yin2017syntactic,rabinovich2017abstract,hayati2018retrieval,iyer2018mapping,hashimoto2018edit},
these programs
are mostly evaluated on syntactic metrics
(e.g., exact match and BLEU score)
rather than functional correctness.
In contrast, program synthesis
in the programming languages community emphasizes
producing programs with the correct semantics,
typically captured by a set of input-output test cases
that the program must compute correctly
\cite{gulwani2011automating,feser2015synthesizing}.
However, input-output pairs usually give little information
about the intermediate states of the program,
making it difficult to synthesize long programs.

Synthesizing a general class of programs of significant length and internal complexity is extremely challenging without some description of the steps of computation.
To that end, we propose a framework %
for synthesizing programs
from natural language \emph{pseudocode} and
\emph{test cases}.
The test cases provide the semantic specification, while the pseudocode provides guidance for the intermediate computations
the program should perform.    

To synthesize a functionally correct program,
instead of relying on the top-one translation
of the pseudocode,
we
search over the space of possible translations
to find one that passes the test cases.
In this work,
we treat the translation of each pseudocode line
as a discrete portion of the program.
As illustrated in Figure~\ref{fig:running-ex},
each pseudocode line translates
to a code line with approximately
one or two atomic statements,
and a program can be synthesized
by choosing a candidate translation
for each pseudocode line.

However,
common search methods for machine translation,
such as beam search over the possible
sequences of code tokens \cite{zavershynskyi2018naps},
only use the sparse reward of whether
the program succeeds.
Without a proper credit assignment
to pinpoint the causes of program failures,
it is difficult to guide search toward
more promising programs. %
Since empirically 88.7\% of failures during search
are due to compilation errors,
we propose to perform credit assignment
based on signals extracted from compilation results.
At a high level,
when a program fails to compile,
we use an \emph{error localization method}
to identify which portion of the program
is responsible for the failure,
and then focus the search on alternative translations
of the pseudocode for that portion.
We propose two error localization methods. %
The first method uses a multiclass classifier
to pick one of the code lines as the offending line,
which is then down-weighted in subsequent search iterations.
In contrast to previous error correction models
\cite{gupta2017deepfix},
our model also uses the error message and pseudocode
for prediction.
This is crucial when the compilation error can be fixed
in multiple ways,
but only some of which are consistent with the pseudocode.
The second method, prefix-based pruning, %
uses additional compilations
to find a code prefix that causes the error.
Unlike the classification model, the identified code prefix
is guaranteed to be erroneous and can be blacklisted entirely. 

For evaluation, we collected and release a new dataset,
SPoC (Search-based Pseudocode to Code)\footnote{
The dataset can be downloaded at \url{https://cs.stanford.edu/~sumith/spoc/}.
},
containing 18,356 C++ programs (14.7 lines on average).
In contrast to other language-to-code datasets
\cite{ling2016latent,oda2015learning,iyer2018mapping},
all programs contain multiple test cases for validation.
And in contrast to the closely related NAPS dataset
\cite{zavershynskyi2018naps},
which also contains test cases but only 6\% human-authored pseudocode,
all programs in SPoC are associated with
human-authored pseudocode
of a consistent annotation granularity.
\refsec{dataset} details the comparison
between SPoC and related datasets.

Using the top-one translation of the pseudocode
yields a %
success rate of 24.6\%
on the test set.
Under a limited budget of 100 synthesis trials (i.e.,
100 code compilations and executions),
our best method achieves a success rate of 44.7\%.
The multiclass error localization model
reduces the number of synthesis trials needed
in 15.5\% of the programs,
with a median absolute reduction of 26 trails
and a median relative reduction of 42\%.
On the other hand,
prefix-based pruning slightly increases
the number of compilations for easier problems,
but is more effective on harder programs,
making it outperform the multiclass classifier
under larger budgets.

\begin{figure}[t]
\centering
{
\small
\begin{tabular}{@{}rl@{\hspace{-.2em}}l@{}}
$i$ & \multicolumn{1}{c}{$x_i$} 
& \multicolumn{1}{c}{$y_i$} \\
\midrule
1 & \pseu{in function main}
  & \verb|int main() {| \\
2 & \verb|  |\pseu{let n be integer}
  & \verb|  |\verb|int n;| \\
3 & \verb|  |\pseu{read n}
  & \verb|  |\verb|cin >> n;| \\
4 & \verb|  |\pseu{let A be vector of integers}
  & \verb|  |\verb|vector<int> A;| \\
5 & \verb|  |\pseu{set size of A = n}
  & \verb|  |\verb|A.resize(n);| \\
6 & \verb|  |\pseu{read n elements into A}
  & \verb|  |\verb|for(int i = 0; i < A.size(); i++) cin >> A[i];| \\
7 & \verb|  |\pseu{for all elements in A}
  & \verb|  |\verb|for(int i = 0; i < A.size(); i++) { | \\
8 & \verb|    |\pseu{set min\_i to i}
  & \verb|    |\verb|int min_i = i;| \\
9 & \verb|    |\pseu{for j = i + 1 to size of A exclusive}
  & \verb|    |\verb|for(int j = i+1; j < A.size(); j++) {| \\
10 & \verb|      |\pseu{set min\_i to j if A[min\_i] > A[j]}
  & \verb|      |\verb|if(A[min_i] > A[j]) { min_i = j; }| \\
11 & \verb|    |\pseu{swap A[i], A[min\_i]}
  & \verb|    |\verb|swap(A[i], A[min_i]);| \\
12 & \verb|  |\pseu{print all elements of A} 
  & \verb|  |\verb|for(int i=0; i<A.size(); i++) cout<<A[i]<<" ";| \\
& & \verb|}|
\end{tabular}
\vspace{.5em}

\begin{tabular}{llcll}
\textbf{Public test case 1 (out of 5):} & 
\verb|5 3 2 4 1 5| & $\to$ & \verb|1 2 3 4 5| \\
\textbf{Hidden test case 1 (out of 8):} & 
\verb|8 9 2 4 5 6 2 7 1| & $\to$ & \verb|1 2 2 4 5 6 7 9 | \\
\end{tabular}
}
\caption{
Given $L$ pseudocode lines $x_{1:L}$
(with indentation levels $\ell_{1:L}$)
and public test cases,
our task is to synthesize a program
with code lines
$y_{1:L}$.
The program is evaluated against both
public and hidden test cases.
}
\label{fig:running-ex}
\end{figure}

\section{Problem statement}

\reffig{running-ex} illustrates the setup of our synthesis task. 
The system is given
(a) a sequence $x$ of $L$ \emph{pseudocode lines}
$x_1, x_2, \dots, x_L$,
where each $x_i$ is a string with indentation level $\ell_i$;
and (b) $k$ \emph{public test cases}
in the form of input-output string %
pairs
$(\tin_1, \tout_1), \dots, (\tin_k, \tout_k)$.
The task is to synthesize a program $y$ consisting of
$L$ \emph{code lines} $y_1, y_2, \dots, y_L$.
The program is \emph{accepted} if it successfully compiles
and passes all public test cases
(i.e., the compiled binary prints the string $\tout_i$
after reading the input $\tin_i$) %
as well as $k'$ additional \emph{hidden test cases}
$(\htin_1, \htout_1), \dots, (\htin_{k'}, \htout_{k'})$.

At training time, the system has access to
a training dataset where each example contains
pseudocode $x$, a gold program $y$,
and both public and hidden test cases.

At test time, the system has access to
pseudocode $x$, public test cases (not hidden ones),
and a computation budget.
For a fair comparison
under different computing environments,
we use the number of \emph{synthesis trials} as the budget,
where in each trial, the system can issue a single call
to the compiler and execute the compiled program
on public test cases.
The system must output a single final program,
which will be validated on both public and hidden test cases.

\section{Dataset}
\label{sec:dataset}

Recall that our goal is to synthesize programs 
of significant length and complexity. 
To this end, we argue that it is important to have 
both description of the intermediate computation
and a functional specification.
Table~\ref{tab:comparison} shows that 
most existing datasets %
\cite{ling2016latent,oda2015learning,iyer2018mapping}
have some varying levels of description, but
lack mechanisms to validate the correctness of programs.
This inevitably leads previous work to resort to proxy metrics, 
such as exact match accuracy, BLEU score, and tree node F1 score,
which only measure syntactic similarity
rather than functional correctness
\cite{ling2016latent,yin2017syntactic,rabinovich2017abstract,hayati2018retrieval,iyer2018mapping,hashimoto2018edit}.

One notable exception and the inspiration for our work 
is the NAPS dataset \cite{zavershynskyi2018naps}
which contains both description (pseudocode) 
and a functional specification (test cases) 
of competitive programming problems.
However, most of their pseudocode is generated
by heuristic rule-based templates,
which in turn has a low information content
compared to the human-authored pseudocode.
Furthermore, the dataset suffers from the inconsistent
granularity of text description, as 
the artificial pseudocode is fine-grained
(e.g., ``increase var0 by 1'')
whereas the human-written pseudocode
tends to be abstract (e.g., ``compute factorial'')
as the annotators were encouraged to provide
high-level descriptions. 
This discrepancy is reflected on the ratio of the length of pseudocode to that of code, which is 1:1.82 in their synthetic dataset, and 1:3.26 in their human-authored dataset.

As no existing dataset contains both high-quality
human-authored description with a consistent level of granularity
and a mechanism to validate functional correctness, %
we created a new dataset called \ours~(Search-based Pseudocode to Code),
which consists of programs, pseudocode, and test cases.
The programs are non-trivial solutions to competitive programming problems,
and each program is paired with public and hidden test cases.
We collected natural language pseudocode for each code line
from curated crowdworkers, which by design, ensures the consistent granularity of description.

\begin{table}[t]
\small
\centering
\caption{Datasets for %
natural language 
to code. In contrast to other datasets,
our SPoC dataset contains human-authored pseudocode
with a consistent granularity of description
and test cases.
}
\begin{tabular}{@{}lcccccc@{}}
\toprule
 & \mtg & \hearthstoneshort & \django & \concode\footnotemark[1] & \naps\footnotemark[2] & SPoC \\
 & \cite{ling2016latent}
 & \cite{ling2016latent}
 & \cite{oda2015learning,ling2016latent}
 & \cite{iyer2018mapping}
 & \cite{zavershynskyi2018naps}
 & \\
\midrule
Programming language & Java & Python & Python & Java & UAST & C++ \\
Number of programs (total) & 13,297 & 665 & 18,805 & 2,184,310 & 17,477 & 18,356 \\
Lines per program (average) & 30.4 & 7.7 & 1 & 4.4 & 21.7 & 14.7 \\
\midrule
Type of natural language input & \multicolumn{2}{c}{--- card text ---} & comment & documentation & \multicolumn{2}{c}{--- pseudocode ---} \\
Additional input & \multicolumn{2}{c}{--- card metadata ---} & - & class context & - & - \\
Granularity of text description & program & program & line & program & varies & line \\
& (class) & (class) &  & (method) &  &  \\
Fraction of human-annotated text & 100\% & 100\% & 100\% & 100\% & 6\% & 100\% \\
Number of annotators (total) & n/a & n/a & 1 & n/a & n/a & 59 \\
\midrule
Test cases & \xmark & \xmark & \xmark & \xmark & \cmark & \cmark \\
Number of test cases (average) & - & - & - & - & 7.5 & 38.6 \\
\bottomrule
\end{tabular}

\label{tab:comparison}
\end{table}
\footnotetext[1]{We counted the number of programs in the released dataset. %
Since the programs are provided as a sequence of tokens, the number of lines per program is approximated based on the number of \texttt{;}, \texttt{\{}, and \texttt{\}}.}
\footnotetext[2]{We excluded partial programs (smaller pieces of full programs) in the dataset when counting.}

\subsection{Data collection}

\paragraph{Programs and test cases.}
Similar to the NAPS dataset \cite{zavershynskyi2018naps},
we scraped competitive programming \emph{problems}
and their test cases
from \url{codeforces.com}.
Each problem
has multiple \textit{programs}
submitted by participants
as solutions to the problem.
We collected accepted C++ programs from
problems 
marked as the easiest level based on their metric. %
Based on our pilot study,
we filtered out programs
with constructs that are difficult to consistently annotate
with pseudocode
(i.e., programs with \verb|#define| macros, classes, 
structs, templates, switch statements, and mallocs).
\paragraph{Decomposition.}
We decompose each program into code lines.
To obtain slightly higher-level descriptions
for common constructs,
we group any block with only one statement with the preceding
control statement
(e.g., the one-line for loop
``\verb|for (int i = 0; i < n; i++) cin >> x[i];|''
allows a high-level description
``read n values into x'').

\paragraph{Pseudocode.}
We recruited 59 crowdworkers on Amazon Mechanical Turk 
to write pseudocode for each line of code.
To our surprise, we were able to identify 
the workers (rather than curated specialists)
who are capable of annotating C++ code
by using a qualification round,
in which we manually inspected their initial annotations.

\paragraph{Statistics.} Our dataset contains 18,356 programs %
submitted for 677 programming problems.
Each problem has roughly 27 programs,
which are likely to have similar semantics yet different code syntax.
Excluding closing braces and the common ``\verb|int main()|'' line,
each program contains an average of 14.7 lines
(with the minimum of 1
and maximum of 457 lines of code).
The average length of code lines is 9.08 tokens, while
the average length of pseudocode lines is 7.86 tokens.

\paragraph{Training and test sets.}

To evaluate the generalization
on unseen problems and annotation styles,
we created two test sets.
We generated the first test set \testp
by splitting based on problems:
we held out 158 problems (23\% out of 677 problems),
which is equivalent to 1,820 programs
(10.1\% of all programs).
The second test set \testw 
is split by workers:
we held out 7 workers
(12\% out of 59 workers),
which is equivalent to 1,752 programs
(9.7\% of all programs,
with 186 programs overlapping with \testp).
We used the remaining data for training and development
(90:10 split).
\section{Base approach}\label{sec:approach}

\begin{figure}
\centering
\includegraphics[height=2.5in]{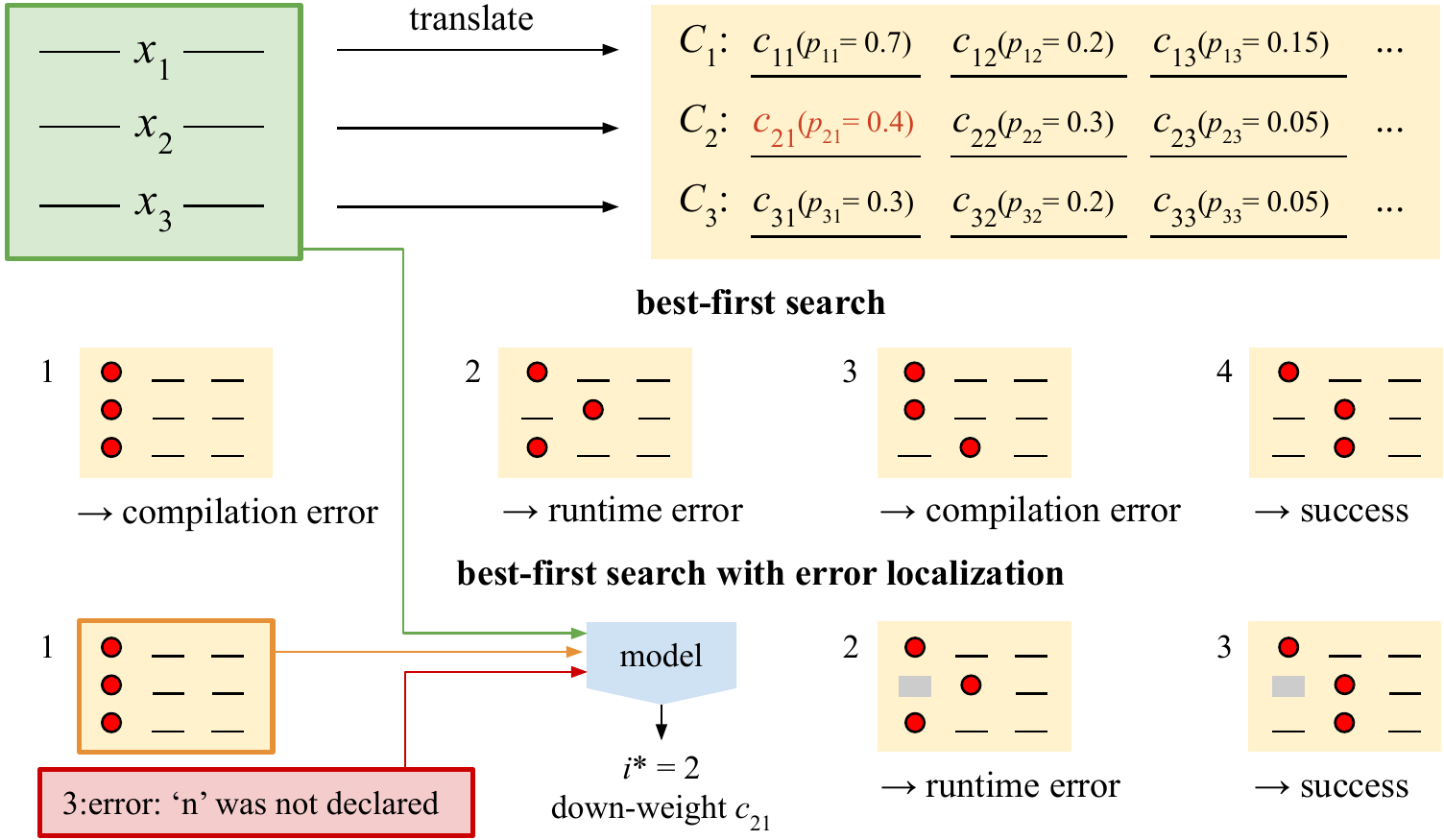}
\caption{
Illustration of best-first search and error localization model. In this example, ($c_{11}, c_{22}, c_{32}$) satisfies the test cases. Best-first search iterates in the order of decreasing probabilities and succeeds in four compiler calls. The %
error localization method
down-weights $c_{21}$, leading to an earlier success.
}
\label{fig:approach}
\end{figure}

As illustrated in Figure~\ref{fig:approach},
our base approach to
synthesizing a program $y_{1:L}$
from pseudocode $x_{1:L}$
and public test cases involves two steps.
First, a translation model
encodes each pseudocode line $x_i$
and generates
$M$ candidate code lines $c_{i1}, \dots, c_{iM}$
to be used as the $i$th code line.
Then, we search over the possible
combinations of candidate translations
until we find a program $\hat y_{1:L}$ that successfully compiles
and passes all public test cases.

\paragraph{Translation.}
To generate candidate code lines,
we use a standard seq2seq translation model
with an LSTM encoder and decoder
\cite{klein2017opennmt}, %
attention-based copying mechanism
\cite{luong2015translation,vinyals2015pointer},
and coverage vector
\cite{tu2016modeling}.
After encoding the pseudocode line $x_i$,
we apply beam search with beam size $M$ to produce
a ranked list of candidates translations
$C_i = (c_{i1}, \dots, c_{iM})$,
where each code line $c_{ij}$ is a sequence of string tokens. %
(We use $M = 100$ for our experiments.)
The model also assigns a probability
$p_{ij} = p(c_{ij} \mid x_i)$ for each candidate $c_{ij}$.
The translation model is trained on pairs
$(x_i, y_i)$ from the training data
using the standard log-likelihood objective.

\paragraph{Best-first search.}
We now describe a basic approach for searching over
the space of possible programs.
Given the candidate lists $C_1, \dots, C_L$,
we can synthesize a program $\hat y$
by picking a candidate $c_{ij[i]}$ from each $C_i$
(where $j[i] \in \{1,\dots,M\}$)
and then concatenate them into a program.
In our search algorithm,
we iterate through programs $\hat y$
in the descending order of probability
$p(\hat y) = \prod p_{ij[i]}$.
To do so,
we maintain a heap of
the combinations $\hat y = (c_{1j[1]}, \dots, c_{Lj[L]})$
indexed by $p(\hat y)$.
The heap initially contains the program
$(c_{11}, \dots, c_{L1})$,
which is the top-one translation of the pseudocode.
In each iteration,
we pop a program $(c_{1j[1]}, \dots, c_{Lj[L]})$
from the heap and test it.
If the program fails
(either from a compilation error, a runtime error,
or a mismatch between the actual and expected test outputs),
we push modified programs
$(c_{1j[1]}, \dots, c_{i(j[i] + 1)}, \dots, c_{Lj[L]})$
for all $i\in\{1,\dots,L\}$
that have not been explored
to the heap.
We continue searching until we either find a program
that passes all \emph{public} test cases
or fully utilize the computation budget.

\section{Error localization}

So far, we have been treating program compilation
and execution as a black box that %
only tells whether
a program passes its test cases.
This sparse signal makes the search process less effective.
For instance,
best-first search will keep using
an erroneous candidate $c_{ij}$
if its probability $p_{ij}$ is high.

To speed up search,
we unpack the black box
and extract more detailed search signals.
In this work,
we focus on compilation errors,
which constitute 
88.7\%
of the failure cases
in best-first search.
When a program $\hat y = (c_{1j[1]}, \dots, c_{Lj[L]})$
fails to compile,
the compiler will report error messages
along with the line numbers where the errors occur.
Unfortunately, the reported line numbers
do not always correspond to the actual location
of the mistake
(e.g., the error ``\emph{`n' was not declared in this scope}''
can occur long after the line where \texttt{n}
should be declared according to the pseudocode).
Empirically, the reported line number does not match
the actual incorrect line 21.7\% of the time.

Therefore, we treat the compilation error message
as a noisy signal,
and propose to use an \emph{error localization method}
to infer the actual portion of the code
that causes the error.
As illustrated in \reffig{approach}, %
the error localization method
has access to the pseudocode $x$,
the synthesized code $\hat y$,
and the \emph{first} error message $(i_\err, m_\err)$
from the compiler, where $i_\err$ is a line number
and $m_\err$ is a message string.
It can then either detect the offending code lines
or abstain.
Depending on the method,
we either down-weight or blacklist the
translation candidates in the offending code lines.

We now introduce two error localization methods:
multiclass classification,
which uses a neural model to predict a single offending line;
and prefix-based pruning, which uses additional
calls to the compiler for detecting
an erroneous code prefix.

\paragraph{Multiclass classification.}
We train a neural multiclass classifier
to predict the offending line $i^*$ among the $L$ lines.
Our model is similar to the error correction model in
\cite{gupta2017deepfix}.
For each line $i$,
we embed the tokens of $x_i$, $y_i$, and $m_\err$,
and then use three separate LSTMs to encode the sequences.
We concatenate the final LSTM hidden states
with the positional encoding \cite{vaswani2017attention}
of the line offset
$\Delta i = i_\err - i$,
and then apply
a feedforward network to
produce the \emph{line embedding} of line $i$.
The $L$ line embeddings are then passed through another LSTM,
and the hidden state of each cell $i$
is passed through a feedforward network
to compute the logit for line $i$.
We return the line $i^*$ with the highest probability
(softmax over logits)
if that probability exceeds a threshold
$\beta_\textrm{mul}$ and abstain otherwise.
We use $\beta_\mathrm{mul} = 0.95$ for the experiments.
Given $i^*$,
we down-weight the current translation candidate of the line $i^*$
so that it is used less often in subsequent search iterations.
Concretely,
we multiply the probability $p_{i^*j[i^*]}$
of the current candidate $c_{i^*j[i^*]}$ in line $i^*$
with a constant factor $\alpha < 1$.
As this affects the heap indices,
we rebuild the heap from scratch
(which takes negligible time)
and continue the search,
skipping any program that has already been explored
before the heap rebuild.

To construct a dataset for training the model,
we consider each program
$y = y_{1:L}$ in the synthesis training dataset,
substitute a single line $y_{i^*}$ with
a candidate $c_{i^*j} \in C_{i^*}$
generated from pseudocode line
$x_{i^*}$,
and then collect any modified program $y'$ that produces
a compilation error with
an error message $(i_\err, m_\err)$.
The model is trained to maximize the log-likelihood
of the offending lines $i^*$.

\paragraph{Prefix-based pruning.}
The multiclass classification method does not guarantee that
the predicted line $i^*$
is actually an offending line.
Furthermore, a candidate code line might be offending in
some contexts but not others
(e.g., a variable re-declaration is no longer offending
if the previous declaration no longer exists).
To address these issues,
we propose an alternative that uses additional
compiler calls to find an offending \emph{prefix}
of the program.
Concretely,
when a compilation error occurs,
we use the compiler to
to find the minimum $i^*$ such that the prefix
$(c_{1j[1]}, \dots, c_{i^*j[i^*]})$,
plus closing braces to complete the program,
fails to compile.
Since programs containing that prefix
will also always %
fail (with very rare exceptions),
we can safely \emph{blacklist} the prefix
from future search iterations.

Each additional compiler call
is counted as one trial toward the synthesis budget.
To save the budget,
we only test $i^* = i_\err - \Delta i$
where $\Delta i \in \{0, 1, 2\}$ corresponds to the three most
frequent offsets. If we fail to find an offending prefix,
we simply abstain.

\section{Experiments}

Our main evaluation metric is \emph{success rate at $B$}:
the fraction of test examples
where the system generates an accepted program
under the budget of $B$ trials.
For error localization methods,
we also consider the reduction in the number of trials used
compared to normal best-first search.

\paragraph{Translation accuracy.}
When evaluating the translation model,
surface-form metrics such as exact sequence match
and BLEU scores fail to account for functional correctness
of the code.
For instance,
a prediction ``\verb|if (b)|''
is functionally equivalent
to the gold code ``\verb|if (b == true)|''
when \texttt{b} is a boolean.
Hence, we instead evaluate the \emph{functional correctness}
of the translation.
To check if a predicted code line $c_{ij} \in C_i$ is
functionally correct,
we replace the code line $y_i$ in the gold program
with $c_{ij}$ and then verify whether
the program still passes both public and hidden test cases.

\newcommand\percentone{
\begin{tikzpicture}[baseline={([yshift={1ex}]current bounding box.south)}]
\begin{axis}[
  xbar stacked,
  width=2in, height=1.1in,
  bar width=0.16in,
  enlarge y limits=0.6,
  axis x line*=bottom,
  xmin=0, xmax=100,
  xlabel={\small\% of programs},
  axis y line*=none,
  symbolic y coords={testw,testp},
  ytick={testw,testp},
  yticklabels={\small\testw,\small\testp},
  y tick label style={align=center},
  point meta=explicit symbolic,
  nodes near coords
]
\addplot[fill=blue!30] coordinates {(18.22,testp) [0] (32.04,testw) [0]};
\addplot[fill=red!30] coordinates {(23.51,testp) [1] (30.48,testw) [1]};
\addplot[fill=brown!30] coordinates {(18.34,testp) [2] (16.34,testw) [2]};
\addplot[fill=green!30] coordinates {(12.54,testp) [3] (8.81,testw) [3]};
\addplot[fill=yellow!30] coordinates {(27.39,testp) [4+] (12.33,testw) [4+] };
\end{axis}
\end{tikzpicture}
}
\newcommand\percenttwo{
\begin{tikzpicture}[baseline={([yshift={1ex}]current bounding box.south)}]
\begin{axis}[
  xbar stacked,
  width=2in, height=1.1in,
  bar width=0.16in,
  enlarge y limits=0.6,
  axis x line*=bottom,
  xmin=0, xmax=100,
  xlabel={\small\% of programs},
  axis y line*=none,
  symbolic y coords={testw,testp},
  ytick={testw,testp},
  yticklabels={{},{}},
  y tick label style={align=center},
  point meta=explicit symbolic,
  nodes near coords
]
\addplot[fill=blue!30] coordinates {(55.23,testp) [0] (71.44,testw) [0]};
\addplot[fill=red!30] coordinates {(23.96,testp) [1] (18.89,testw) [1]};
\addplot[fill=brown!30] coordinates {(9.90,testp) [2] (5.39,testw) [2]};
\addplot[fill=green!30] coordinates {(10.91,testp) [3+] (4.28,testw) };
\end{axis}
\end{tikzpicture}
}

\begin{figure}
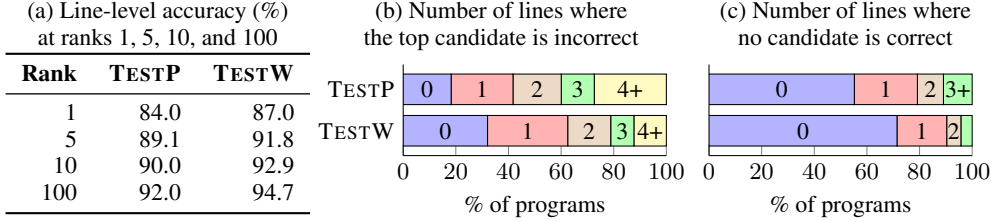

\centering \small
\begin{tabular}{c@{}c@{}c}
{\small (a) Line-level accuracy (\%)} &
{\small (b) Number of lines where} &
{\small (c) Number of lines where} \\
{\small at ranks 1, 5, 10, and 100} &
{\small the top candidate is incorrect} &
{\small no candidate is correct} \\
\begin{tabular}[b]{rrr}\toprule
\textbf{Rank} & \textbf{\testp} & \textbf{\testw} \\ \midrule
1 & 84.0 & 87.0 \\
5 & 89.1 & 91.8 \\
10 & 90.0 & 92.9 \\
100 & 92.0 & 94.7 \\
\bottomrule
\end{tabular} &
\percentone &
\percenttwo 
\end{tabular}
\caption{
(a) While the translation accuracy is high at the line level,
we need to consider the result at the program level.
For each program, we count the number of lines $i$ where
(b) the top candidate $c_{i1}$ is incorrect, and
(c) none of the candidates $c_{ij} \in C_i$ is correct.
}
\label{fig:translation-stats}
\end{figure}

The results in Figure~\ref{fig:translation-stats}(a)
shows that when the lines are considered independently,
the translation model achieves a high accuracy of 84--87\%
under this notion of functional correctness.
However,
the picture is grimmer when we consider the statistics
at the \emph{program} level,
which is what matters for synthesis.
For each program,
we count the number of lines $i$ where
the top candidate $c_{i1}$ is not functionally correct.
Figure~\ref{fig:translation-stats}(b) shows that
only 18.2\% of programs in \testp
and 32.0\% of programs in \testw
have the top candidate correct in every line.
As code lines that are functionally correct in isolation
may be incompatible one another,
the programs formed by combining
the top candidate of each line %
have an even lower success rates of 17.8\% on \testp
and 30.7\% on \testw.

\paragraph{Oracle success rate.}
To compute the maximum achievable success rate given the
lists of candidates,
for each program,
we count the number of lines $i$
where the candidate list $C_i$
does not have any correct candidate.
Figure~\ref{fig:translation-stats}(c) shows that
44.8\% of programs in \testp
and 28.6\% of programs in \testw have
least one difficult line where 
the translation model does not produce a correct prediction
among the top $M = 100$ candidates.
This means a synthesizer with an infinite search budget
would achieve a maximum success rate of 55.2\%
on \testp and 71.4\% on \testw given
our lists of candidates 
(assuming that incorrect candidates do not give a correct behavior when
combined together).
\newcommand{\resultsTestP}{
\begin{tikzpicture}
\begin{axis}[
    width=.5\textwidth,
    height=2in,
    xmin=0, xmax=3000,
    ymin=31, ymax=41,
    xlabel={budget $B$},
    x label style={at={(axis description cs:0.5,0)},anchor=south},
    ylabel={success rate (\%)},
    y label style={at={(axis description cs:0,1)},rotate=-90,anchor=north west},
]
\addplot[mark=none,color=magenta,dashed]
table[x=budget,y=testp-best-first] {figures/results-testp-raw.tsv};
\addplot[mark=none,color=blue]
table[x=budget,y=testp-prefix-pruning] {figures/results-testp-raw.tsv};
\addplot[mark=none,color=green!70!black]
table[x=budget,y=testp-gray-naive] {figures/results-testp-raw.tsv};
\addplot[mark=none,color=red]
table[x=budget,y=testp-gray-advanced] {figures/results-testp-raw.tsv};
\end{axis}
\end{tikzpicture}
}
\newcommand{\resultsTestW}{
\begin{tikzpicture}
\begin{axis}[
    width=.5\textwidth,
    height=2in,
    xmin=0, xmax=3000,
    ymin=51, ymax=61,
    xlabel={budget $B$},
    x label style={at={(axis description cs:0.5,0)},anchor=south},
    ylabel={success rate (\%)},
    y label style={at={(axis description cs:0,1)},rotate=-90,anchor=north west},
]
\addplot[mark=none,color=magenta,dashed]
table[x=budget,y=testw-best-first] {figures/results-testw-raw.tsv};
\addplot[mark=none,color=blue]
table[x=budget,y=testw-prefix-pruning] {figures/results-testw-raw.tsv};
\addplot[mark=none,color=green!70!black]
table[x=budget,y=testw-gray-naive] {figures/results-testw-raw.tsv};
\addplot[mark=none,color=red]
table[x=budget,y=testw-gray-advanced] {figures/results-testw-raw.tsv};
\end{axis}
\end{tikzpicture}
}

\begin{figure}[t]
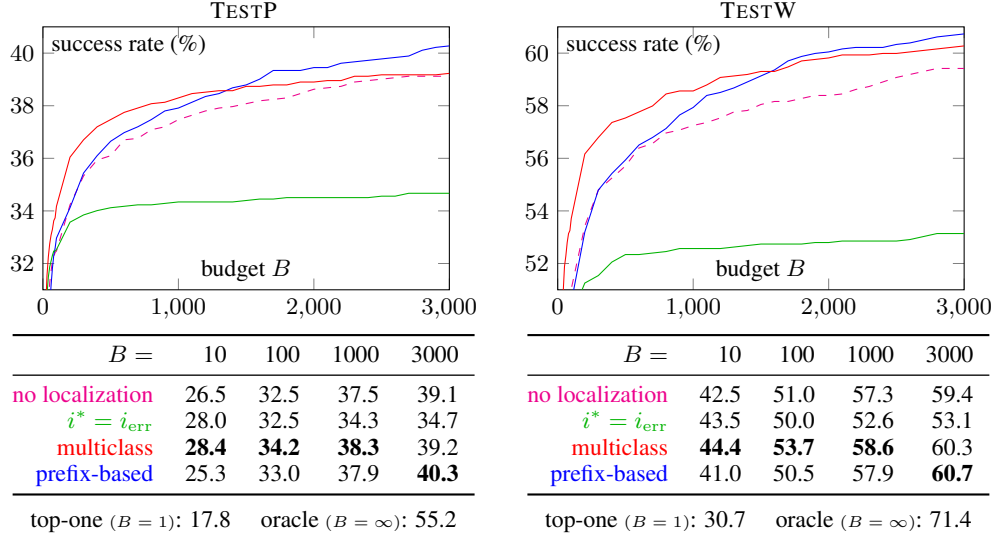

\centering \small
\begin{tabular}{cc}
\testp & \testw \\
\resultsTestP & \resultsTestW \\
\begin{tabular}{@{}rrrrrrr@{}} \toprule
$B=$ & 10 & 100 & 1000 & 3000 \\ \midrule
\color{magenta}{no localization}	&	26.5	&	32.5	&	37.5	&	39.1	\\
\color{green!70!black}{$i^* = i_\err$}	&	28.0	&	32.5	&	34.3	&	34.7	\\
\color{red}{multiclass}	&	\textbf{28.4}	&	\textbf{34.2}	&	\textbf{38.3}	&	39.2	\\
\color{blue}{prefix-based}	&	25.3	&	33.0	&	37.9	&	\textbf{40.3}	\\
\bottomrule
\end{tabular} &
\begin{tabular}{@{}rrrrrrr@{}} \toprule
$B=$ & 10 & 100 & 1000 & 3000 \\ \midrule
\color{magenta}{no localization}	&	42.5	&	51.0	&	57.3	&	59.4	\\
\color{green!70!black}{$i^* = i_\err$}	&	43.5	&	50.0	&	52.6	&	53.1	\\
 \color{red}{multiclass}	&	\textbf{44.4}	&	\textbf{53.7}	&	\textbf{58.6}	&	60.3	\\
\color{blue}{prefix-based}	&	41.0	&	50.5	&	57.9	&	\textbf{60.7}	\\
\bottomrule
\end{tabular} \\ \addlinespace
top-one {\tiny$(B=1)$}: 17.8 \quad
oracle {\tiny$(B=\infty)$}: 55.2 &
top-one {\tiny$(B=1)$}: 30.7 \quad
oracle {\tiny$(B=\infty)$}: 71.4 
\end{tabular}
\caption{Success rates at budgets $B$
of best-first search with different error localization
methods.
}
\label{fig:main-result}
\end{figure}

\begin{table}[t] \centering
\caption{Effects of using error localization methods
on all test examples.
}\small
\begin{tabular}{llrrrrrr} \toprule
& \multicolumn{2}{r}{number of trials:}
& \multicolumn{2}{c}{absolute difference} %
& \multicolumn{2}{c}{relative difference} \\
\cmidrule(r){4-5}
\cmidrule{6-7}
\textbf{method}
& \textbf{effect compared to best-first}
& \textbf{count}
& \textbf{mean}
& \textbf{median}
& \textbf{geo.mean}
& \textbf{median} \\ \midrule
multiclass
& improves number of trials & 13.5 \% & --199.5 & --26.0 & $\times$0.39 & $\times$0.58 \\
& failed to synthesize $\to$ succeeds & 2.0 \% \\ \cmidrule{2-7}
& worsens number of trials & 0.4 \% & +407.5 & +123.0 & $\times$6.70 & $\times$7.04 \\
& succeeded $\to$ fails to synthesize& 1.6 \% \\ \midrule
prefix-based
& improves number of trials & 4.1 \% & --272.6 & --91.0 & $\times$0.45 & $\times$0.57 \\
& failed to synthesize $\to$ succeeds & 1.5 \% \\ \cmidrule{2-7}
& worsens number of trials & 15.7 \% & +68.4 &  +12.0 & $\times$1.65 & $\times$1.63\\
& succeeded $\to$ fails to synthesize & 0.3 \% \\
\bottomrule
\end{tabular}
\label{tab:delta-b}
\end{table}

\paragraph{Synthesis results.}
\reffig{main-result} compares the success rates of
best-first search with and without error localization.
As a baseline,
we try down-weighting the reported error line ($i^* = i_\err$)
whenever a compilation error occurs.
Due to the mismatch between the actual offending line
and the reported line,
the synthesis result deteriorates.
Up to the compilation budget of around $B = 1500$,
the multiclass classifier
improves the success rate the most.
Prefix-based pruning achieves better success rates
for higher budgets,
but since it uses additional compilation calls,
it performs worse under tighter budgets.

\reftab{delta-b} details how the error localization methods
affect the synthesis outcome.
The multiclass model decreases the number of trials
in 15.5\% of all examples,
but since its predictions are not verified,
the model is also more prone to catastrophic failures.
Prefix-based pruning uses additional compilation calls
to verify its verdict,
and thus slightly worsens the number of compilations needed
in a large number of examples.
However, for more difficult programs, the benefit outweighs
the cost.

\begin{figure}[t]
\centering
\newcommand{\yay}[1]{\color{gray}#1}
{
\small
\begin{tabular}{rrlrrl}
(1) && \dots & (2) && \dots \\
&& \yay{let s be a string}
&&& \yay{create int \,{\color{red}l}, p and q} \\
& 7 & \verb|string s ;|
&& 2 & \verb|int |{\color{red}\verb|a|}\verb| , p , q ;| \\
&& \yay{read s}
&&& \yay{read l, p and q} \\
& 8 & \verb|cin >> s ;|
&& 3 & \verb|cin >> l >> p >> q ;| \\
&& \yay{if {\color{red}s is half}}
&&& \yay{print l * p / (p + q)} \\
& 9 & \verb|if ( |{\color{red}\verb|s / 2 == 0|}\verb| )| 
&& 4 & \verb|cout << l * p / ( p + q ) << endl ;|\\
&& \dots &&& \dots
\end{tabular}
}
\caption{
Examples of programs synthesized during search.
In Program 1, prefix-based pruning detects that the prefix
up to line 9 is offending.
In Program 2, the multiclass model incorrectly predicts
line 3 as the offending line, which ultimately leads to a failure.
}
\label{fig:error-ex}
\end{figure}

\paragraph{Error analysis.}
To understand the behavior of error localization methods,
we analyzed several examples from the development data.
Some prototypical examples are shown in \reffig{error-ex}.
Program 1 shows how error localization can improve search.
The condition ``s is half'' in line 9
should be translated as ``\verb|s == "half"|'',
but was instead interpreted as ``\verb|s / 2 == 0|''
and ``\texttt{s \% 2 == 0}''
with high probability,
and hence best-first search spends a significant
amount of budget (1511 trails)
using these incorrect candidates
in the combination.
In contrast, prefix-based pruning
detects them as offending candidates and succeeds earlier
(413 trials).

In contrast, Program 2 shows how an incorrect error localization
can lead to a catastrophic failure.
Here, the multiclass model reads the error message
$m_\err = $ ``\emph{‘l’ was not declared in this scope}''
with line number $i_\err = 3$,
and incorrectly predicts that line 3 is an offending line.
This causes the search to ultimately fail
whereas best-first search finds a correct program
in 80 search iterations.

\section{Related work and discussion}

\paragraph{Program synthesis.} Program synthesis using test cases has been extensively studied
in the programming languages literature.
The most focused and effective method is to formulate %
synthesis as a constraint satisfaction problem \cite{solar2006combinatorial, tate2009equality}, 
which requires
that the synthesis problem can be translated to a theory with effective constraint solvers.
For other problems, brute force enumeration of programs
(with some optimization) works surprisingly well
\cite{massalin1987superoptimizer,bansal2006automatic}, 
but when the search space is too large for enumeration, 
randomized search guided by a cost function can be effective \cite{schkufza2013stochastic}.  
Some works combine aspects of multiple approaches (e.g., \cite{jha2010oracle}). 
For program specifications, the norm is to use input-output pairs. %
However, most synthesized programs are relatively short,
and works that consistently synthesize longer programs
are in the domains where the intermediate computation
is easier to recover from input and output,
such as string
\cite{gulwani2011automating, parisotto2017sql, devlin2017robustfill}
and data structure transformations \cite{feser2015synthesizing,yaghmazadeh2016hierarchy}.
For other domains,
while input-output examples are \emph{precise} in evaluating
functional correctness,
they provide mostly \emph{global} signals and 
inform very little about the intermediate computation,
thereby requiring other forms of specification 
along with input-output examples (e.g., \cite{feng2017api,shi2019frangel}).

\paragraph{Semantic parsing.}
Works on translating natural language specifications
to executable programs, as discussed in \refsec{dataset},
are closely related to semantic parsing 
whose goal is to map natural language utterances to formal representation. 
One of its traditional tasks is to parse a given question 
(usually a single sentence) into %
an executable database query
\cite{zelle96geoquery,zettlemoyer07relaxed,liang11dcs,berant2013freebase,zhong2017seq2sql,yaghmazadeh2017sqlizer}. 
Instead of a single query, some work aims to
parse a sequence of utterances into queries 
that can be sequentially executed \cite{andreas2015alignment,iyyer2016answering,long2016projections}.
However, the sequences are still relatively short (e.g., maximum 5 sentences).

\paragraph{Error localization.} 
Error localization in the context of automated program repair
has been an active topic of research.
Many recent work that uses neural models to localize errors
has focused on localizing and correcting syntax errors \cite{gupta2017deepfix}
or a class of well-defined semantic errors such as variable misuse and variable replace
\cite{allamanis2018varmisuse,devlin2017varreplace,vasic2019neural}.
Other work identifies error locations by interpreting compiler error messages
\cite{hartmann2010what,bhatia2018synfix}.
Likewise, our multiclass error localization model uses
compilation errors to locate offending code;
however, since the code is tied to pseudocode,
we also use the signal from pseudocode
to distinguish ambiguous cases %
(e.g., in Program 2 of \reffig{error-ex},
while changing either line 2 or line 3 can fix the error,
a correct model should choose line 2 as the offending line
with respect to the pseudocode.)

\subsubsection*{Acknowledgements}
We thank Shivam Garg, Jason Koenig, Nadia Polikarpova, Alex Polozov and Rishabh Singh for valuable feedback at different stages of the project. This work was supported by NSF grant CCF-1409813, NSF CAREER Award IIS-1552635 and a grant from Amazon.

\bibliographystyle{abbrv}
\bibliography{all}

\end{document}